\documentclass[a4paper,10pt,conference]{IEEEtran}

\usepackage{graphicx}
\usepackage{latexsym}

\usepackage{array}
\usepackage{rotating}
\usepackage{listings}
\usepackage{pdflscape}
\usepackage{booktabs}

\usepackage{todonotes}
\usepackage{balance}

\begin{document}

\title{Predicting and visualizing psychological attributions with a deep neural network}

\author{\IEEEauthorblockN{$^{\ast}$Edward Grant\IEEEauthorrefmark{1},
$^{\ast}$Stephan Sahm\IEEEauthorrefmark{2}, $^{\ast}$Mariam Zabihi\IEEEauthorrefmark{3} and
Marcel van Gerven\IEEEauthorrefmark{4}}
\IEEEauthorblockA{Radboud University\\
Nijmegen, The Netherlands\\
Email: \IEEEauthorrefmark{1}edward339@gmail.com,
\IEEEauthorrefmark{2}stephan.sahm@gmx.de,
\IEEEauthorrefmark{3}mariam.zabihi@gmail.com,
\IEEEauthorrefmark{4}m.vangerven@donders.ru.nl}\\\textit{$^{\ast}$Denotes equal contribution}}

\maketitle

\begin{abstract}
Judgments about personality based on facial appearance are strong effectors in social decision making, and are known to have impact on areas from presidential elections to jury decisions. 
Recent work has shown that it is possible to predict perception of memorability, trustworthiness, intelligence and other attributes in human face images. The most successful of these approaches require face images expertly annotated with key facial landmarks. We demonstrate a Convolutional Neural Network (CNN) model that is able to perform the same task without the need for landmark features, thereby greatly increasing efficiency. The model has high accuracy, surpassing human-level performance in some cases. Furthermore, we use a deconvolutional approach to visualize important features for perception of 22 attributes and demonstrate a new method for separately visualizing positive and negative features.
\end{abstract}

\IEEEpeerreviewmaketitle

\section{Introduction}

Facial attributions for intelligence, attractiveness, dominance and trustworthiness have been shown to exhibit a strong effect on social decision making, with far-reaching consequences from choosing between presidential candidates to jury decisions in criminal legal cases \cite{ballew2007predicting,re2013looking}.

Despite this, reliably predicting how a face will be perceived has proven to be difficult. Some of the best current methods require images hand-annotated with facial landmark features, which is time consuming \cite{ICCV13_Khosla}.

Given the success of CNNs in image recognition tasks \cite{krizhevsky2012imagenet}, the CNN model is a natural choice for visual attribution of psychological characteristics. In addition to superior performance in vision tasks, CNNs are able to learn visual features from image data and do not require additional human input or hand-crafted features.

Previous work on modelling attribute perception was conducted by Khosla et al.\ who introduced a method for characterizing face images using key facial points, histogram information, SIFT features and hand-annotated landmark facial features \cite{ICCV13_Khosla}. Using these features, it was possible to accurately predict attributions for many psychological and demographic attributes. In contrast, we use a CNN to learn features directly from RGB images without the need for facial landmark annotations, which are time consuming and can introduce bias. Using these learned features the CNN is able to predict attribution labels with high accuracy, surpassing human-level performance in some cases. In addition we demonstrate a method to visualize general attribution features learned by the CNN.

One important distinction is between the \textit{perception} of psychological attributes (attributions) and other tests for an attribute. Visual perception has important social consequences, but is not always a good indicator of more robust measurements for an attribute. For example, Rezlecu et al.\ showed that perceived trustworthiness, is not significantly correlated with measured trustworthiness \cite{rezlescu2012unfakeable}. In contrast Kleiser et al.\ showed that perceived intelligence is associated with measured intelligence in men but not women \cite{kleisner2014perceived}. In this experiment we focus solely on \textit{perception} of attributes.

Several methods exist for visualizing the features learned by a neural network. These methods can broadly be divided into approaches that require a target image to be forward propagated through the network before the activity of a target feature detector can be projected back into image space \cite{DBLP:journals/corr/ZeilerF13,  DBLP:journals/corr/MahendranV14} and image-free approaches that generate an image that maximizes a class score \cite{ DBLP:journals/corr/SimonyanVZ13, nguyen-2015-CVPR-deep-neural-networks}.

The first type of approach has the benefit of visualizing features from a real example image: the second approach can be more general because it does not rely on a single image example. We use the first approach, but visualize the mean of many examples, thus retaining both the benefit of visualizing features from real images and the generality of an image-free approach. This is only possible because the images we used contain faces that have roughly the same pose. If this was not the case the features could become obscured by each other.

To accomplish feature visualization we use the deconvnet proposed by Zeiler and Fergus \cite{DBLP:journals/corr/ZeilerF13}.  In this approach, a target image is forward propagated through the network, and all activations except the targets are set to zero. The target activation is projected back into image space by passing the activation back through the network using deconvolution and a special kind of up-sampling. The projected image contains the image features most responsible for the target activation.

The resulting visualization represents all of the features important for an attribute. However, by manipulating the final layer, network weights we are able to show that these features can be decomposed into their positive and negative components. Using this method we separately visualize both the positive and negative features for attributions. 

\section{Methods}

\begin{figure*}[!htbp]
\centering
\includegraphics[scale=0.6]{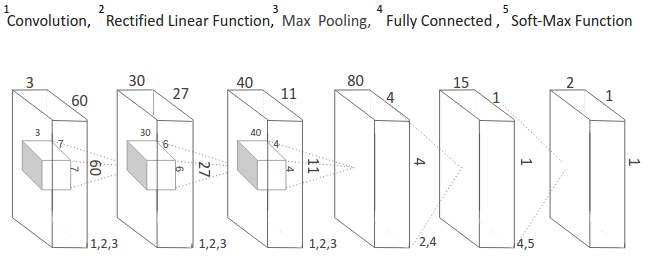}
\caption{CNN schematic for attribution prediction. The network takes as input an image and outputs a binary class prediction. Within the network each convolutional layer transforms the output of the previous layer using learned filters that are convolved across overlapping sub-regions. The output of each layer is subject to one or more of the following nonlinear transformations: rectified linearity, max pooling and softmax.}
\label{fig:CNNNet}
\end{figure*}

\subsection{Dataset}
Our example set comprised the annotated subset of the 10K face data\-base collected by Bainbridge et al \cite{bainbridge2013intrinsic}. This set consists of 2222 face photographs as RGB images annotated with psychological and demographic labels. These data were collected in the form of a rating from 15 participants for psychological features and 13 participants for demographic features.

\subsection{Preprocessing}
Each image was labelled as belonging to one of two equally sized classes for each attribute based on its ratings for a target attribute. For gender, the ratings were binary already and the classes were balanced by removing images randomly from the larger class. Images were square cropped and resized to $60\times60$ pixels. Images were zero-centered by subtracting the mean pixel value from all images.

\begin{table*}[t]
\centering
\caption{Accuracy and correlations for psychological and demographic attributions. Used abbreviations: acc - accuracies; corr - correlations. Values in \textbf{bold} stand for significant improvement compared to the respective human performance on $5$\% alpha level. Attributions in \textbf{bold} denote significant improvement in both accuracy and correlation (again compared to human performance).}
\begin{tabular}{@{}lccccc@{}}
\toprule
attributions  & acc CNN & acc Human & acc SVM & corr CNN & corr Human \\ \midrule
age           & 0.75    & 0.98      & 0.65    & 0.58     & 0.86       \\
attractive    & 0.74    & 0.72      & 0.59    & \textbf{0.59}     & 0.47       \\
calm          & 0.63    & 0.65      & 0.56    & 0.30     & 0.22       \\
caring        & 0.77    & 0.75      & 0.63    & \textbf{0.65}     & 0.46       \\
common        & 0.64    & 0.66      & 0.54    & \textbf{0.34}     & 0.13       \\
confident     & 0.69    & 0.71      & 0.55    & \textbf{0.44}     & 0.31       \\
egotistic     & 0.70    & 0.66      & 0.61    & \textbf{0.51}     & 0.29       \\
\textbf{emotional}     & \textbf{0.73}    & 0.64      & 0.58    & \textbf{0.51}     & 0.19       \\
emotStable    & 0.69    & 0.68      & 0.57    & \textbf{0.47}     & 0.29       \\
familiar      & 0.63    & 0.59      & 0.53    & 0.30     & 0.12       \\
\textbf{friendly}      & \textbf{0.80}    & 0.77      & 0.64    & \textbf{0.68}     & 0.53       \\
gender        & 0.94    & 1.0      & 0.73    & 0.89     & 0.98       \\
\textbf{happy}         & \textbf{0.84}    & 0.78      & 0.65    & \textbf{0.73}     & 0.59       \\
intelligent   & 0.64    & 0.68      & 0.56    & 0.39     & 0.27       \\
interesting   & 0.68    & 0.67      & 0.57    & \textbf{0.43}     & 0.22       \\
\textbf{kind}          & \textbf{0.78}    & 0.74      & 0.65    & \textbf{0.66}     & 0.47       \\
memorable     & 0.63    & 0.64      & 0.55    & 0.29     & 0.16       \\
responsible   & 0.70    & 0.71      & 0.57    & \textbf{0.50}     & 0.35       \\
\textbf{sociable}      & \textbf{0.79}    & 0.75      & 0.64    & \textbf{0.67}     & 0.47       \\
trustworthy   & 0.75    & 0.71      & 0.62    & \textbf{0.61}     & 0.38       \\
typical       & 0.64    & 0.66       & 0.53    & 0.30     & 0.15       \\
weird         & 0.65    & 0.66      & 0.53    & 0.41     & 0.29       \\ \bottomrule
\end{tabular}
\label{tab:results}
\end{table*}

\subsection{Training the Network}
For each attribution, each of $11$ CNN models were trained within one fold of an $11$-fold cross validation setting. Each example was represented exactly once in the test set.  The network comprised of three convolutional layers, two fully connected layers and a softmax layer. 

Training was performed using stochastic gradient descent with momentum set at $0.9$, a batch size of $60$, a learning rate of $0.005$ and weight decay of $0.001$. The models were trained using MatConvNet \cite{vedaldi15matconvnet} on a Tesla K80 GPU. See Figure~\ref{fig:CNNNet} for a more detailed description of the network structure. 

\subsection{Performance Measures}
Two performance measures were computed. The outputs from the CNN are denoted by the probability of an image belonging to the positive or the negative class. These values were thresholded at $50\%$ and afterwards compared to the true binarization of the image set (individually for each attribution). The fraction of correct predictions was used to determine accuracy.
As a baseline, a linear support vector machine (SVM) was trained on the same data, and corresponding accuracy measurements were obtained. Furthermore, single human accuracy was computed using a leave-one-out strategy over the given (binarized) dataset.

Correlation values refer to standard correlation coefficients. They were computed between the CNN output probabilities and the continuous human assessment from the dataset. Again, single human correlation was obtained by a leave-one-out procedure over the (continuous) dataset for a respective attribution. While accuracy values show the correctness of the predictions, the correlations show how the variation within the predictions mimics the true variation in the data.

\subsection{Statistics}
CNN performance was tested for significance in two respects: whether it was significantly different to human performance and an a random baseline. 

The CNN performance values were revealed to be consistent enough over trials in order to reasonably approximate them as being constant. We therefore assume that training the CNN twice will result in exactly the same prediction output -- no randomness is involved. Hence, to test for the null hypothesis that the CNN performance could be produced by baseline or humans, we only need to estimate how probable it is that the same or higher performance is generated by the baseline or humans respectively. For this the respective distributions were approximate.
As we are interested in significant \emph{improvements}, the hypothesis test was one-sided, and the critical value for an alpha set to $5\%$. For higher values, we reject the null hypothesis and concluded that CNN performance is significantly better than baseline or human performance.

The accuracy measure as defined above counts the fraction of correct binary predictions. Hence, the random baseline is the mean of a respective number of independent coin-flips. As there are $2222$ images, the baseline distribution is the mean of $2222$ independent fair Bernoulli distributed random variables.
The distribution of the correlations is less simple to derive. A completely random CNN would output an arbitrary probability prediction between $0$ and $1$, independently for each image. This corresponds to a vector of $2222$ independent standard uniform random variables. The correlation measurement then demands computation of the correlation between the CNN response and each attribution dataset individually. Instead of deriving an analytic expression for such a correlation of a random vector with a given data vector of size $2222$, it is easier to simulate these correlation values. For this study, $100,000$ random samples were generated from a uniform random vector, and each correlated with all $22$ attribution datasets. 

For comparison with human performance, the distributions of human performance was approximated. For both accuracy and correlations, a bootstrap estimation procedure was used. A short explanation of the bootstrap approach follows. Our dataset was generated by a small number of participants. Estimating the distribution of human performance over the whole population would mean taking a new (random) subset of participants and computing the same performance measure again, and again and again, thereby simulating its distribution. As this is obviously not feasible, the method known as bootstrap approximation regards the given subset as the population itself. Instead of taking a new random subset of the whole world population, a random sub-subset of the given subset is chosen and the human performance measurement computed. For this study, sampling with replacement was used and again $100,000$ samples were generated.

\subsection{Visualizing Features Using Deconvolution}
Attribution features were visualized using a deconvnet \cite{DBLP:journals/corr/ZeilerF13}. Using this method, the target image was first forward propagated through the network and the location with maximum activation for all pools stored. 
The activation for the target attribute was passed back through the network using deconvolution. At each pooling layer an approximate inverse to the pooling operation was performed by up-sampling the image and placing the feature map pixel values at the location with maximum activation previously stored during pooling.

The target activation is caused by positive and negative predictions from the previous layer, and so deconvolution visualizes an image with both positive and negative features. In order to separately visualize features that positively and negatively contribute to an attribution prediction, we set the negative weights in the final layer to zero to visualize the positive features and set the positive weights to zero to visualize the negative features. This is possible because the final layer nodes' inputs are the weighted sum of the previous layer activations, which are all positive or zero because of the rectified linear activation function. By setting the positive or negative weights to zero in the final layer, the effect of the positive or negative features on the prediction is isolated. Using this method, we create three visualizations for each attribution. One contains all the features responsible for a prediction, one only negative features and one contains only positive features. 

The deconvolution approach is typically used to visualize the features of a single example image. Because most images contain faces in roughly the same pose, we visualized the mean of all feature representations for each attribute. This results in an image with the mean features for an attribution. This process was repeated for positive and negative features.  


\section{Results}

\begin{table*}[p] 
\centering
\caption{
Deconvolved Features. Per attribution, the dataset was binarized into low versus high values (class 0 and 1 respectively) and a deep neural network was trained for classification. Following training, two additional network variants were created retaining only the positive(+) or negative(-) final layer weights. This allows for separate visualization of positive and negative features for each attribution. The images displayed are the averaged deconvolution results over the whole test dataset. For instance, for the attribution gender a pronounced mouth region is common for class 0 (female) while nose and eyebrows are important features for class 1 (male).
}
\label{tab:deconv}
\begin{tabular}{
    >{\centering\arraybackslash}p{1.8cm}
    >{\raggedright\arraybackslash}p{1.44cm}
    >{\raggedright\arraybackslash}p{1.44cm}
    >{\raggedright\arraybackslash}p{3.0cm}
    >{\centering\arraybackslash}p{1.8cm}
    >{\raggedright\arraybackslash}p{1.44cm}
    >{\raggedright\arraybackslash}p{1.44cm}
    >{\raggedright\arraybackslash}p{1.44cm}
    }
\toprule
Attribution & Features & Features(-) & Features(+) & Attribution & Features & Features(-) & Features(+)\\
\midrule

~~~age{\newline}(young/old) & \parbox[c]{1em}{\includegraphics[scale=0.88]{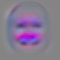}} &
\parbox[c]{1em}{\includegraphics[scale=0.88]{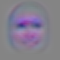}} & \parbox[c]{1em}{\includegraphics[scale=0.88]{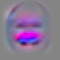}} &

attractive & \parbox[c]{1em}{\includegraphics[scale=0.88]{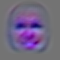}} & \parbox[c]{1em}{\includegraphics[scale=0.88]{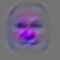}} & \parbox[c]{1em}{\includegraphics[scale=0.88]{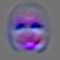}}\\

calm & \parbox[c]{1em}{\includegraphics[scale=0.88]{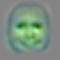}} & \parbox[c]{1em}{\includegraphics[scale=0.88]{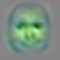}} & \parbox[c]{1em}{\includegraphics[scale=0.88]{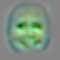}} &

caring & \parbox[c]{1em}{\includegraphics[scale=0.88]{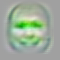}} & \parbox[c]{1em}{\includegraphics[scale=0.88]{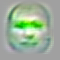}} & \parbox[c]{1em}{\includegraphics[scale=0.88]{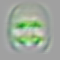}}\\

common & \parbox[c]{1em}{\includegraphics[scale=0.88]{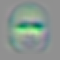}} & \parbox[c]{1em}{\includegraphics[scale=0.88]{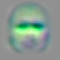}} & \parbox[c]{1em}{\includegraphics[scale=0.88]{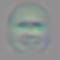}} &

confident & \parbox[c]{1em}{\includegraphics[scale=0.88]{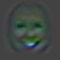}} & \parbox[c]{1em}{\includegraphics[scale=0.88]{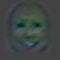}} & \parbox[c]{1em}{\includegraphics[scale=0.88]{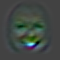}}\\

egotistic & \parbox[c]{1em}{\includegraphics[scale=0.88]{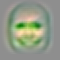}} & \parbox[c]{1em}{\includegraphics[scale=0.88]{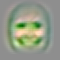}} & \parbox[c]{1em}{\includegraphics[scale=0.88]{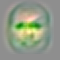}} &

emotional & \parbox[c]{1em}{\includegraphics[scale=0.88]{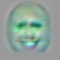}} & \parbox[c]{1em}{\includegraphics[scale=0.88]{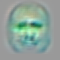}} & \parbox[c]{1em}{\includegraphics[scale=0.88]{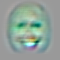}}\\

emotStable & \parbox[c]{1em}{\includegraphics[scale=0.88]{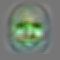}} & \parbox[c]{1em}{\includegraphics[scale=0.88]{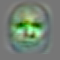}} & \parbox[c]{1em}{\includegraphics[scale=0.88]{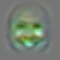}} &

familiar & \parbox[c]{1em}{\includegraphics[scale=0.88]{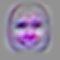}} & \parbox[c]{1em}{\includegraphics[scale=0.88]{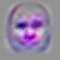}} & \parbox[c]{1em}{\includegraphics[scale=0.88]{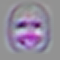}}\\

friendly & \parbox[c]{1em}{\includegraphics[scale=0.88]{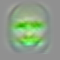}} & \parbox[c]{1em}{\includegraphics[scale=0.88]{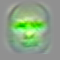}} & \parbox[c]{1em}{\includegraphics[scale=0.88]{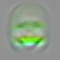}} &
~~gender{\newline}(female/male) & \parbox[c]{1em}{\includegraphics[scale=0.88]{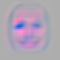}} &
\parbox[c]{1em}{\includegraphics[scale=0.88]{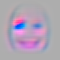}} & \parbox[c]{1em}{\includegraphics[scale=0.88]{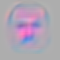}}\\

happy & \parbox[c]{1em}{\includegraphics[scale=0.88]{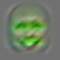}} & \parbox[c]{1em}{\includegraphics[scale=0.88]{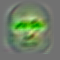}} & \parbox[c]{1em}{\includegraphics[scale=0.88]{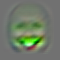}} &

intelligent & \parbox[c]{1em}{\includegraphics[scale=0.88]{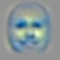}} & \parbox[c]{1em}{\includegraphics[scale=0.88]{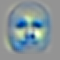}} & \parbox[c]{1em}{\includegraphics[scale=0.88]{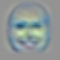}}\\

interesting & \parbox[c]{1em}{\includegraphics[scale=0.88]{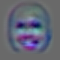}} & \parbox[c]{1em}{\includegraphics[scale=0.88]{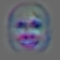}} & \parbox[c]{1em}{\includegraphics[scale=0.88]{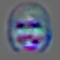}} &

kind & \parbox[c]{1em}{\includegraphics[scale=0.88]{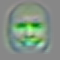}} & \parbox[c]{1em}{\includegraphics[scale=0.88]{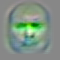}} & \parbox[c]{1em}{\includegraphics[scale=0.88]{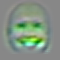}}\\

memorable & \parbox[c]{1em}{\includegraphics[scale=0.88]{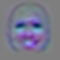}} & \parbox[c]{1em}{\includegraphics[scale=0.88]{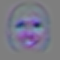}} & \parbox[c]{1em}{\includegraphics[scale=0.88]{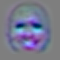}} &

responsible & \parbox[c]{1em}{\includegraphics[scale=0.88]{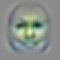}} & \parbox[c]{1em}{\includegraphics[scale=0.88]{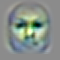}} & \parbox[c]{1em}{\includegraphics[scale=0.88]{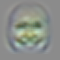}}\\

sociable & \parbox[c]{1em}{\includegraphics[scale=0.88]{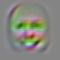}} & \parbox[c]{1em}{\includegraphics[scale=0.88]{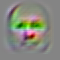}} & \parbox[c]{1em}{\includegraphics[scale=0.88]{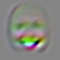}} &

trustworthy & \parbox[c]{1em}{\includegraphics[scale=0.88]{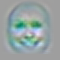}} & \parbox[c]{1em}{\includegraphics[scale=0.88]{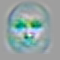}} & \parbox[c]{1em}{\includegraphics[scale=0.88]{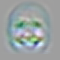}}\\

typical & \parbox[c]{1em}{\includegraphics[scale=0.88]{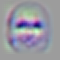}} & \parbox[c]{1em}{\includegraphics[scale=0.88]{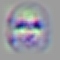}} & \parbox[c]{1em}{\includegraphics[scale=0.88]{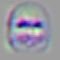}} &

weird & \parbox[c]{1em}{\includegraphics[scale=0.88]{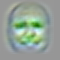}} & \parbox[c]{1em}{\includegraphics[scale=0.88]{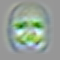}} & \parbox[c]{1em}{\includegraphics[scale=0.88]{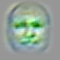}}\\

\bottomrule
\end{tabular}
\end{table*}

\subsection{Performance of the CNN, Human and SVM Classifiers}
The CNN accuracy was higher than SVM accuracy in all cases. The averaged CNN accuracy over all attributions was $71.86$\% with standard deviation $0.080$. For correlations, the mean was $0.511$ and the standard deviation $0.163$.

CNN accuracy as well as correlations between CNN predictions and human assessments were significantly better than chance for all attributions. Compared to human correlations, a significant improvement was found for $14$ attributions: attractive, caring, common, confident, egotistic, emotional, emotional stability, friendly, happy, interesting, kind, responsible, sociable and trustworthy. Compared to human accuracies, the CNN was significantly better in predicting $5$ attributions, namely emotional, friendly, happy, kind and sociable. See Table~\ref{tab:results} for a summary.

\subsection{Visualization of CNN Features Involved in Attribution}

Table~\ref{tab:deconv} visualizes the CNN features involved in attribution. Here we identified a number of salient properties of the visualizations.

As expected the CNN positive features for a happy attribution look very much like a smile. In contrast the negative features focus around the eyes and a down-turned mouth. These two feature sets compete with each other to produce a prediction for happy.

Some of the attribution features we visualize have been studied before and we find a general coherence between previous findings and the CNN feature visualizations. 

The CNN features important for gender discrimination centered around the eyes and lips. This coheres with existing evidence that gender perception can be modulated by subtly changing the color of the lips and eyes in a gender-neutral image \cite{russell2009sex,stephen2010lip}.

Todorov et al.\ show that there is a significant correlation between human facial features and perceived trustworthiness. These features are mainly located at the center of the face, including the eyebrows, cheekbones, nose and chin. The shape of these features generates a spectrum of trustworthiness impressions. A longer narrower nose indicated increased perceived trustworthiness. In addition, the shape of the cheekbones was found to be important \cite{todorov2008evaluating}. The features for trustworthiness shown in Table~\ref{tab:deconv} show that the nose and cheekbones are also important features for CNN attribution of trustworthiness.  

We further observed striking differences in color hue and saturation between features for many attributes. Important features for age, gender and attractiveness are found mainly in the red and blue channels of the target image. The green channel contains important features for friendly and happy, and other attributes have features with a combination of colors. Features for sociable are mainly found in the red channel in the nose area and in the green channel around the eyes and mouth. Although it is interesting to see the color of the features learned by the network, it is advisable not to over-interpret these findings. Just because the network learns features in a specific color channel does not necessarily mean that similar features cannot be found in other channels. Further work is need to determine the relevance of color in attribute perception, for example by comparing the predictive performance of the present model with that of a model trained only on gray-scale images.



\section{Discussion}

This work shows that deep neural networks can be used to accurately predict rated personality traits from face images, even surpassing human-level performance in some cases. 

The coherence between CNN features for attribution and features found to be used by humans is interesting but not unexpected. CNNs and humans both exploit common structures in natural images to perform vision. In addition, CNNs are loosely inspired by real neural structures and, similar to CNNs, there is strong evidence that the human visual system is organized in a hierarchy of feature detectors of increasing complexity \cite{gucclu2015deep}.

By visualizing positive and negative features separately, we showed that for some attributions all features are important, whereas for others, the positive and negative features are better indicators of the existence of an attribution. For example, the negative features for responsible are much more pronounced, whereas for memorable the positive features are more important (see Table~\ref{tab:deconv}).

Considering the performance of the CNN compared to human performance, it is interesting that the correlation values are far more often significantly outperformed by CNN than accuracy values are. The CNN can replicate the variability within the assessments better than the overall classification. We used classification rather than regression to allow for binary feature representations using deconvolution, but a similar network trained using regression may yield superior performance for attribution accuracy. 

Many of the positive and negative features for attributions discriminate between innate features such as face shape or distance between the eyes. Gender and age are good examples of such attributions. In contrast, for some attributions positive and negative features appear to be discriminated by expressions. Good examples of these attributions are confident, friendly, happy, kind and emotional. Smiling is a positive feature for each of these attributions, suggesting that perception of these attributions can be changed through facial expression, unlike gender and age. Other attributes contain a mixture of fixed and expressive features. Attractiveness appears to be influenced by both face shape and the orientation of the lips. A smile is a feature of attractiveness, suggesting that perception of attractiveness can be modulated through expression. 

In conclusion, we have shown that CNNs can be used to predict rated psychological and demographic attributions and to analyze the visual features that contribute to the prediction of these attributions. This can have practical applications as well as providing new insights into the psychological underpinnings of personality ratings.

\section*{Acknowledgment}
We would like to thank Wilma Bainbridge for the Human Faces 10k dataset and Umut G\"u\c{c}l\"u for providing the deconvnet code.



%



\end{document}